\documentclass[numbers,sort]{article}

\usepackage{graphicx}
\usepackage{booktabs}
\usepackage{tabularx}
\usepackage{amsmath}

\usepackage[final]{neurips_2024}

\usepackage[utf8]{inputenc} 
\usepackage[T1]{fontenc}    

\usepackage[colorlinks,
            linkcolor=blue,       
            anchorcolor=blue,
            citecolor=blue, 
            ]{hyperref}      
\usepackage{url}            
\usepackage{booktabs}       
\usepackage{amsfonts}       
\usepackage{nicefrac}       
\usepackage{microtype}      
\usepackage{xcolor}         

\title{Imagine with the Teacher: Complete Shape in a Multi-View Distillation Way}

\author{
  Zhanpeng Luo$^{1}$\ \quad Linna Wang$^{1}$ \quad Guangwu Qian$^{1}$ \quad Li Lu$^{1}$\\
  $^1$Computer College, Sichuan University \\
  \texttt{\{luozhanpeng,lenawang\}@stu.scu.edu.cn} \\
  \texttt{\{g\_qian,luli\}@stu.scu.edu.cn} \\
}

\begin{document}
\maketitle

\begin{abstract}
Point cloud completion aims to recover the completed 3D shape of an object from its partial observation caused by occlusion, sensor’s limitation, noise, etc. When some key semantic information is lost in the incomplete point cloud, the neural network needs to infer the missing part based on the input information. Intuitively we would apply an autoencoder architecture to solve this kind of problem, which take the incomplete point cloud as input and is supervised by the ground truth. This process that develops model’s imagination from incomplete shape to complete shape is done automatically in the latent space. But the knowledge for mapping from incomplete to complete still remains dark and could be further explored. Motivated by the knowledge distillation’s teacher-student learning strategy, we design a knowledge transfer way for completing 3d shape. In this work, we propose a novel View Distillation Point Completion Network (VD-PCN), which solve the completion problem by a multi-view distillation way. The design methodology fully leverages the orderliness of 2d pixels, flexibleness of 2d processing and powerfulness of 2d network. Extensive evaluations on PCN, ShapeNet55, and MVP datasets confirm the effectiveness of our design and knowledge transfer strategy, both quantitatively and qualitatively. Committed to facilitate ongoing research, we will make our code publicly available.

\end{abstract}

\section{Introduction}

 Point cloud completion plays an essential role in 3D vision applications and remains an active research topic in recent years. Throughout this period, a diverse range of learning-based approaches have been introduced, many of which have shown promising results. As research progresses, incremental improvements in result accuracy and model efficiency have been reported. \cite{yuan2018pcn,xie2020grnet,yu2021pointr,xiang2021snowflakenet,chen2023anchorformer}, Despite these advancements, the inherent sparsity and significant structural incompleteness of captured point clouds continue to impede the efficacy of existing models in generating fully satisfactory outcomes\cite{zhu2023svdformer}. We observe two intriguing problems in two fields in the point cloud processing and raise two questions about the current points completion task.

In the field of point cloud classification, the success of SimpleView surprise us\cite{goyal2021revisiting,yan2022let}, which prove the efficacy and robustness of a 2d network in the point cloud processing. Even simple multi-view CNN with the depth map of projected point cloud, can achieve on par or better results than sophisticated state-of-the-art methods on ModelNet40. Looking back to the completion task, while the common solution is to use pointnet++\cite{qi2017pointnet++} structure to extract information directly from the input point cloud\cite{xiang2021snowflakenet,zhu2023svdformer,zhou2022seedformer}, we are wondering if there is still potential for further enhancement? Since a multi-view CNN could do better on the classification problem, we are naturally infer that a CNN-based encoder could get a more descriptive representation for the incomplete input. This prompts the question:\textbf{ Can we reproduce the Simpleview’s success in the completion task?}

In the field of point cloud completion, intuitively we would apply an encoder-decoder structure to solve the completion-like problem implicitly. But recovering an unordered partial point cloud is a challenging task\cite{li2023high}. Existing approaches that rely on decoding a latent feature to recover the complete shape, often lead to the completed point cloud sometimes over-smoothing or losing details on edges. In order to lower the learning difficulty and recover better details, some methods are adopted to exploit the information from both the input and ground truth, such as concatenate the input on the generated point cloud and subsample into the original amount\cite{zhu2023csdn}. And another typical solution is to give the ground truth supervision multiple times, such as a coarse-to-fine paradigm.\cite{wang2020cascaded}  But could we lower the learning difficulty one more step? Would the information from the ground truth be further explored? How can we make the dark knowledge in the latent space more explicit? This raises us the second question:\textbf{ Can we leverage some kind of technique that further lower the learning difficulty in the completion task?}

First, to explore the first question is rather trivial. Following SimpleView\cite{goyal2021revisiting}, we could directly apply a multi-view CNN on a depth image group that the point cloud project to. But a simple shared 2d network for all images do not consider the relation between views. This is especially important for the completion tasks, because sometimes it is very hard for even for human to imagine a complete object from partial observation. Therefore we design a multi-view encoder and a decoder considering the information loss during the CNN pooling operation. Also, since transformation from point cloud to depth map would lose information, we reintroduce the original point cloud information during the decoder stage and use both 2d and 3d feature for point generation. This consists of the basic struture of our model.

Next, to explore a method to lower the learning difficulty, we are inspired by the knowledge distillation technique, the siamese network design, and the knowledge transfer strategy and adopt a teacher–student knowledge transfer strategy to exploit the rich information in the ground truth point cloud. Specifically, the teacher and student share the same structure. The teacher is trained with partial point cloud and complete depth map as input and is supervised by ground truth. After training a teacher who build well-established mapping from complete depth map and partial point cloud to complete point cloud, we froze the teacher and add a feature-level distillation supervision on student. At the student side, we use teacher’s pre-trained weight as initiation, froze all module except the multi-view encoder and train it with a smaller learning rate. The student would not only inherit teacher's semantic understanding but also learn to gradually adapt to the incomplete input and mimic teacher's feature and point prediction. By experiment, we have shown that the knowledge transfer strategy would produce better coarse point and better result.

With the help of this designing methodology and learning strategy, our method outperforms state-of-the-art completion networks on many widely used benchmarks. We would also like to summarize our contribution: 
We devise a new method for points completion achieving current state-of-the-art result on many widely-used dataset by solving two problems we observed. We also extend a new knowledge transfer training diagram on the point cloud completion task, which can be easily adopted for the related problems. We are committed to releasing code for reproducibility and future research.

\section{Related Work}
\paragraph{Point cloud completion}

Like traditional methods for point cloud learning early attempts at 3D shape completion usually adopt intermediate aids (i.e., voxel, grids) \cite{xie2020grnet, dai2018scancomplete}. After the success of Point Net, the community put more attention on the point-based method. Recently, due to some new technique such as attention, and large model, research introduce multi-modality to the completion process, such as paired image\cite{zhang2021view}, RGBD image\cite{aiello2022cross} and text\cite{kasten2024point}. Within the completion model design, common solution is an encoder decoder architecture in a coarse-to-fine paradigm\cite{wang2020cascaded}. A set of coarse points usually first predicted from the input information. During each decoding stage, the coarse point would be upsampled and is supervised by the ground truth. SnowFlakeNe\cite{xiang2021snowflakenet} summarize this as a “Feature Extraction – Seed Generation – Point Generation” paradigm. Recent work put lots of effort on the decoding part, or the point generation part. FBNet\cite{yan2022fbnet} adopts the feedback mechanism during refinement and generates points in a recurrent manner. LAKe-Net\cite{tang2022lake} integrates its surface-skeleton representation into the refinement stage. SVDFormer\cite{zhu2023svdformer} identify a structure similarity between each stage.  These improvements often would either increase the running time or increase model complexity. Also, other methods improve on the loss function chamfer distance\cite{lin2024infocd,lin2023hyperbolic} to help better reconstruct shape.

\paragraph{Multi-View 3d Understanding}
Early works in 3d shape learning usually adopt multi-view projection or 3D voxelization to transform the irregular point clouds into regular representations\cite{qi2016volumetric}. Before the invention of pointnet, we have to operate 2d/3d CNN on these regular representations. But recently, SimpleView\cite{goyal2021revisiting} has drawn people’s attention again to the usage of 2d network in 3d understanding. SimpleView shows us a simple multi-view CNN could perform on par or better than other sophisticated point-based method in the classification problem with great cross-dataset generalization.  Lots of work have been exploring the usage of multi-view projection in other 3d understanding task. Tackling 3D vision tasks with this indirect approach has two main advantages: (i) mature and transferable 2D computer vision models (CNNs\cite{ronneberger2015u}, Transformers\cite{dosovitskiy2020image} etc.), and (ii) large and diverse labeled image datasets for pre-training (e.g., ImageNet\cite{deng2009imagenet}). Furthermore, recent advances in differentiable rendering allow for end-to-end deep learning pipelines that render multi-view images of the 3D data and process the images by CNNs/transformers/diffusion\cite{ho2020denoising} to obtain a more descriptive representation for the 3D data. 

\paragraph{Knowledge Distillation}
Knowledge distillation \cite{hinton2015distilling}, is first proposed as a model compression method, which pushes the student network to mimic the soft logits of the teacher network in the orginally setting. However, there are few works introducing this technique directly to the point cloud analysis.  The image based feature-level distillation would cause negative transfer because either point-feature alignment or point-pixel alignment fails for data’s irregularity and sparsity\cite{zhang2023pointdistiller,yan2022let}. Some methods provide a workaround for the problem. PointDistiller\cite{zhang2023pointdistiller} apply the local distillation and the reweighted learning strategy , which cluster the local neighbouring points and extract information with dynamic graph convolution.  ProtoTransfer\cite{tang2023prototransfer} use a class-wise prototype distillation method to align student in a cross modality way. Let image give you more\cite{yan2022let} choose to align teacher and student on the point cloud loss, rather than a feature level alignment. In our work, we utilize the regularity of pixels and can directly align teacher’s and student’s feature to transfer the knowledge.

\section{Methodology}
Here, we present the task formulation and an overview of VD-PCN.
\begin{figure}
  \centering
  \includegraphics[width=\textwidth]{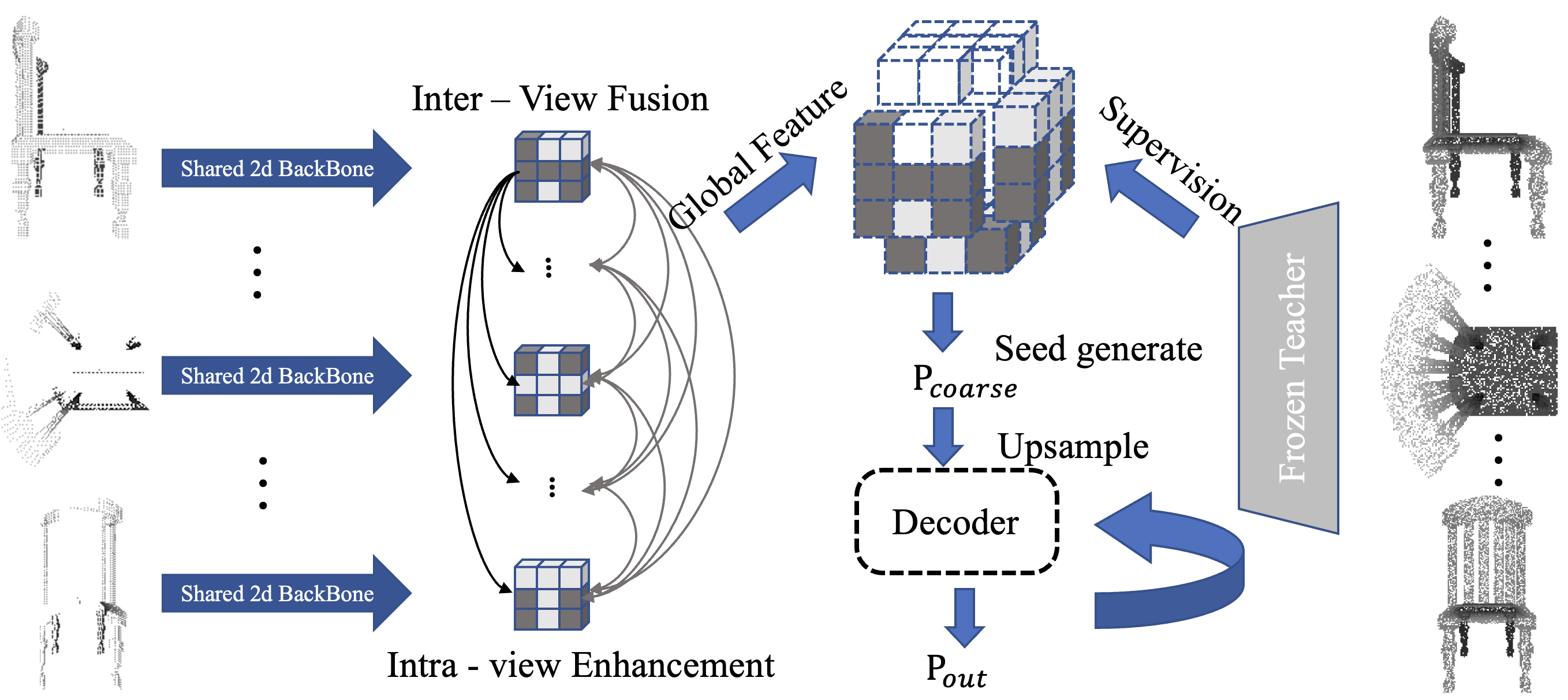}
  \caption{Model Overview}
\end{figure}

\subsection{General Pipeline}  Given a set incomplete unordered points $P_{in} \in \mathbb{R}^{N_{in} * 3}$, the model need to recover a complete point set $P_{out} \in \mathbb{R}^{N_{out} * 3}$. SnowFlakeNet\cite{xiang2021snowflakenet} has formulated this task into three stages: feature extraction, seed generation and point upsampling. In the feature extraction stage, input point cloud is encoded into a global feature, which would be further used to generate a coarse point cloud $P_{coarse} \in \mathbb{R}^{N_{coarse} * 3}$.  as seed. Then in the upsampling stage, this coarse point cloud would finally be upsampled into the same resolution as $P_{out}$.

Applying a 2d network to solve the completion problem, we first takes depth maps for the input point cloud from $k$ viewpoints and get an image group, represented as $ I = \{I_i | i = 1, \ldots, k\} $, $I_{i} \in \mathbb{R}^{1 \times H \times W}$.  Then $I$ would be sent into a shared 2d backbone, getting a multi-view feature group $F = \{ F_i | i = 1, \ldots, k \}$. To further equip our model with a panoramic view, we design a multi-view encoder to aggregate information from each viewpoint. Then the global feature was sent to generate seed and unsample until high resolution. 

When the knowledge transfer strategy is applied, we reformulate the working pipeline as a "Multi-View Perceptual Encoder, Distilled Knowledge Transfer, and Dual Modality Decoder" process. The teacher and student share the same structure. Firstly, a model would be trained with partial point cloud and complete depth map and would be supervised by the ground truth. After the teacher build a well-established mapping from complete depth map to complete point cloud, we froze the teacher and apply a feature-level distillation supervision. Also, we initiate student with teacher's weight. and train with a smaller learning rate and epoch.

\subsection{Model Design}
As we describe in the general pipleline, VD-PCN consist of three part, the multi-view encoder, distilled knowledge transfer and dual-modality decoder. 

\paragraph{Multi-View Perceptual Encoder}

As we discussed above, VD-PCN is initiated with a 2d backbone processing the $k$ view depth images. We choose the left branch of a classical Unet\cite{ronneberger2015u} as our 2d backbone for its larger receptive field. And we do not choose the attention based vision transformer since we hope to build a model in a comparable parameter scale with point-based method. Processed by a shared 2d network, the $k$ views are reshaped into the batch size dimension, sent to the Unet, and downsampled into a multi-view feature map $F_v \in \mathbb{R}^{k \times C \times H_1 \times W_1}$. 

To enable our model with a global awareness of all the $k$ views on one object, we design a multi-view encoder to aggregate each view's information. Inspired by Leap\cite{jiang2023leap}, we choose to use a cross attention on $F_v$ to achieve a panoramic awareness for the object.  Firstly, to find the relationship between distinct views, we design an Intra-View Fusion (IVF) layer, which do cross attention with $f_{0}$ as query and $f_{1...k}$ as key \& value and update $f_{0}$ into $f'_{0}$  This could be formulate as:
\[
f'_{0} = FFN(CrossAttention(K = f_{0}, Q\&V= f_{1...k-1} ))
\]

Following the global integration facilitated by the IVF Layer, the Intra-View Enhancement (IVE) Layer focuses on refining the feature maps within each view. Utilizing self-attention mechanisms, the IVE Layer enhances the relation within the view group. For the IVE layer, we use a self-attention mechanism where the key, query, and value are the image features group $F$. It can be formulated as:

\[
 F'_{v} = FFN(SelfAttention(Q\&K\&V = F_{v} ))
\]

where $F_v'$ stands for the updated value for the whole feature group. And $F_v$ is the original feature group with $f_{0}$ updated. To maximize the effectiveness of our encoder, the IVF and IVE Layers undergo multiple iterations of refinement. This iterative process allows for the progressive enhancement of the global feature map. The updated feature map $F'_v \in \mathbb{R}^{k \times C \times H_1 \times W_1}$ would be further max pooled into a global feature tensor $F'_g \in \mathbb{R}^{k \times C }$.

\paragraph{Distilled Knowledge Transfer}
Another core part of VD-PCN is the knowledge transfer distillation strategy. Initially, we train a teacher who share the same structure with student, which takes complete depth map and partial point cloud as input. The teacher is supervised by the ground truth. After building a well-established mapping from complete depth map to complete point cloud, we begin to train our model with a feature-level supervision by the teacher. During the experiment, we find that downsample the ground truth to make depth images to would produce better result. A reasonable guess is fewer points already contain enough information for completion\cite{wu2024fsc} and a small amount of points would prevent object's structure occlusion and bring less noise. 

We do not adopt the online knowledge distillation\cite{guo2020online} to compare in a similar computation cost framework. Therefore, the teacher's multi-view encoder is frozen, takes the complete depth map as input and generate teacher's feature map $ F^{T}_{v} \in \mathbb{R}^{k \times C \times H_1 \times W_1}$ and max-pooled global feature tensor $ F^{T}_{g} \in \mathbb{R}^{k \times C }$. 

Then the student, who take a partial point cloud depth image would mimic the teacher's behaviour. In specific, we align $ F^{S}_{v}$ with $ F^{T}_{v}$ in a $L_2$ loss and $F^{S}_{g}$ with $ F^{T}_{g}$ in a $L_1$ loss.  

\[
\mathcal{L}_{\text{KD}} = \tau_{1} \cdot \sum_s |P^T_v - P^S_{v}|^2 + \tau_{2} \cdot \sum_s |P^T_g - P^S_{g}| 
\]
Here we choose L2 loss for $F_{v}$ and L1 loss for $F_{g}$. By earlier experiments, different type of loss and scale factor $\tau_{1}$ and $\tau_{2}$ seem to achieve similar effect at the end. Therefore we simply set $\tau{1} = \tau{2} = 1$ and introduce two type of loss in the distillation stage.

And here, we also wanted to discuss the distillation  design in the point completion task. As far as we know, we are the first one to design a knowledge distillation method on the point completion task. This is because a naive knowledge distillation method would cause negative transfer for the unorderliness of the point cloud. A feature-level alignment stragety would fail when the teacher's input do not have the same order as student's\cite{zhang2023pointdistiller}. But in VD-PCN, we apply CNN on projection depth images, which leverage the orderliness of pixels to align teacher's and student's feature to achieve a knowledge transfer effect.

\paragraph{Dual-Modality Decoder}
\begin{figure}
  \includegraphics[width=\textwidth]{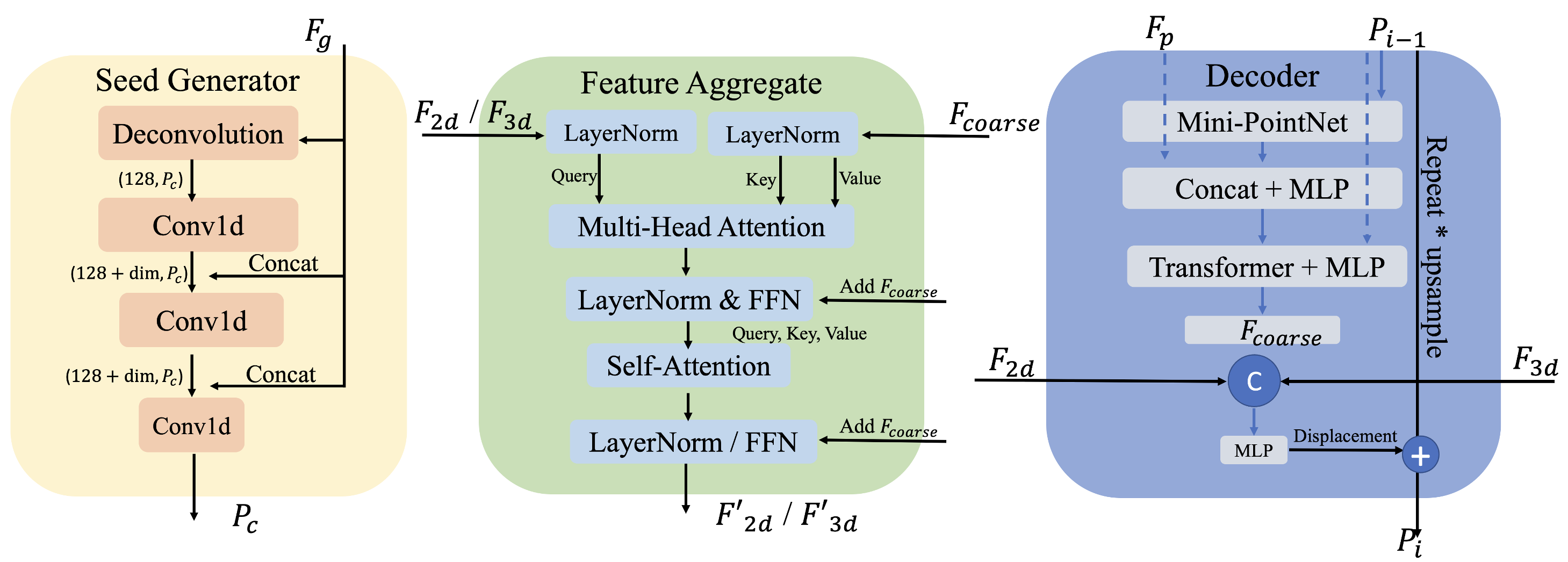}
  \caption{Model detail illustrated for three modules in the decoder part.}
\end{figure}
The decoder generate seed from $F_g$ and upsample in a coarse to fine way until final high-resolution point cloud. Following SnowFlakeNet, we adapt a seed generater, a deconvlution followed by residual connected MLP would generate the first coarse point cloud . And to lower the difficulty to generate seed. A common technique when generating $P_c$ is to concatenate the input on the $P_c$ and subsample it into the same resolution it with $P_c$.

At $i_{th}$ stage, the coarse points $P_{i-1}$ from the previous stage is utilized. A mini-pointnet would first project the coarse into a high-dimensional space, then we concatenate the global feature on these points. After that, we apply a transformer to strengthen the mapping between $F_{coarse}$ and $P_{coarse}$.

At the feature aggregation module, the key ideas in the decoder design is to re-introduce the lost information from two modality. The first information loss is in the transformation from point cloud $P_in$ to multi-view depth images $I$ and the second information loss is the pooling operation from $F_v$ to $F_p$. Therefore we use a feature aggregation module to introduce 2d and 3d information into the upsampling process. In detail, we sent apply a cross attention module with $F_{coarse}$ as query and $F_p$ / $F_v$ as Key \& Value. After feature aggregation, $F_{2d}$ and $F_{3d}$ are concatenated together and sent into a MLP to predict the displacement for the upsampled point cloud. Then we simply repeat $P_{i-1}$ and add the predicted displacement to produce $P_i$.

\subsection{Training Loss}
In our implementation, we use Chamfer distance as the primary loss function. We supervise all predicted point sets $P_c ... P_{n}$ with ground truth $P_{gt}$. And we add the distillation loss $\mathcal{L}_{kd}$. The scale factor is chosen as 1 during our experiment. The training loss can be formulated as:

\[
\mathcal{L} = \mathcal{L}_{CD}(P_c, P_{gt}) + \sum_{n} \mathcal{L}_{CD}(P_i, P_{gt})+ \tau_{0} \cdot \mathcal{L}_{kd}
\]

\section{Experiment}
\subsection{Experiment Setting}
\paragraph{Dataset}We empirically verify the merit of VD-PCN on PCN, ShapeNet-55 and MVP dataset. All of these datasets are created from ShapeNet. We follow the standard protocols into evaluate models on these three datasets. In particular, ShapeNet-55 contains 55 categories and includes 41, 952 and 10, 518 shapes in the training and testing sets, respectively. The evaluations on ShapeNet-55 are conducted on the point cloud data masked with a ratio of 25\%50\% and 75\%, accordingly formulating the completion task at three difficulty levels of simple (S), moderate (M) and hard (H). The MVP\cite{pan2021variational} dataset contain 62400 sample from 2400 CAD models for training and 41600 samples for 1600 CAD models for testing. MVP dataset also provide 4 different resolution (2048, 4096, 9192, 16384). We report the highest resolution in the paper and provide other in the supplementary material.

\paragraph{Implementation Detail} We implement our VD-PCN on the PyTorch platform\cite{paszke2019pytorch}. The number of depth camera viewpoint $k$ is 6, with resolution $\left[ H, W\right] =  224 \times 224$. The 2d backbone Unet downsample the input in 16 times, producing a $\left[ H1, W1\right] =  14 \times 14$ and the channel $ c =  512$. We choose set the number of coarse point $P_c = 128$. During experiment, we find that set $P_c $ to 128 or 256 would produce better performance, comparing to 512 and 1024. We take two stage for coarse to fine point generation. In training process,  we choose the AdamW optimizer\cite{loshchilov2017decoupled} with weight decay = 0.0005. For the teacher, the learning rate is set as 0.0002 and we train it with 200 epoch. For the distillation training, we choose a small learning rate at 0.0001 and train it with 200 epoch. We use the L1 Chamfer Distance as the point cloud training loss, a MSE loss for the teacher / student's global feature alignment and L1 loss for teacher / student's visual feature loss. All the scale factor $\tau_{0}$, $\tau_{1}$, $\tau_{2}$ are set to 1. 

\paragraph{Evaluation metrics}We employ the L1 Chamfer Distance as the evaluation metrics for the PCN and L2 Chamfer Distance for ShapeNet-55, MVP datasets. We also report the common used F score for evalution. On the ShapeNet55, there are three test difficulty and we also report an average Chamfer Distance on table \ref{S3}. 

\subsection{Evaluation on benchmarks}
\begin{table*}[htbp]
  
  \centering
  \caption{Performance comparison in terms of L1 Chamfer Distance $\times 10^3$ (CD$_{L1}$) on the PCN dataset. The Chamfer Distance performances of each category and the averaged result across all categories are all listed. (Lower CD$_{L1}$ is better)}
  \resizebox{\textwidth}{!}{
  \begin{tabular}{l|cccccccc|c}
    \toprule
    Method & Plane & Cabinet & Car & Chair & Lamp & Sofa & Table & Boat & CD$_{L1}$ \\
    \midrule
    FoldingNet\cite{yang2018foldingnet} & 9.49 & 15.80 & 12.61 & 15.55 & 16.41 & 15.97 & 13.65 & 14.99 & 14.31 \\
    PCN\cite{yuan2018pcn}  & 5.50 & 22.70 & 10.63 & 8.70 & 11.00 & 11.34 & 11.68 & 8.59 & 9.64 \\
    GRNet\cite{xie2020grnet} & 6.45 & 10.37 & 9.45 & 9.41 & 7.96 & 10.51 & 8.44 & 8.04 & 8.83 \\
    PoinTr\cite{yu2021pointr} & 4.75 & 10.47 & 8.68 & 9.39 & 7.75 & 10.93 & 7.78 & 7.29 & 8.38 \\
    SnowFlakeNet\cite{xiang2021snowflakenet} & 4.29 & 9.16 & 8.08 & 7.89 & 6.07 & 9.23 & 6.55 & 6.40 & 7.21 \\
    SeedFormer\cite{zhou2022seedformer} & 3.85 & 9.05 & 8.06 & 7.06 & 5.21 & 8.85 & 6.05 & 5.85 & 6.74\\
    AnchorFormer\cite{chen2023anchorformer} & 3.70 & 8.94 & 7.57 & 7.05 & 5.21 & 8.40 & 6.03 & 5.81 & 6.59 \\
    SVDFormer\cite{zhu2023svdformer} & 3.62 & 8.79 & 7.46 & 6.91 & 5.33 & 8.49 & 5.90 & 5.83 & 6.54 \\
    \textbf{Ours}       & \textbf{3.52} & \textbf{8.51} & \textbf{7.36} & \textbf{6.55} & \textbf{5.04} & \textbf{8.27} & \textbf{5.73} & \textbf{5.58} & \textbf{6.32} \\
    \bottomrule
  \end{tabular}
    }
  \label{PCN}
\end{table*}

\begin{figure}
  \centering
  \includegraphics[width=\textwidth]{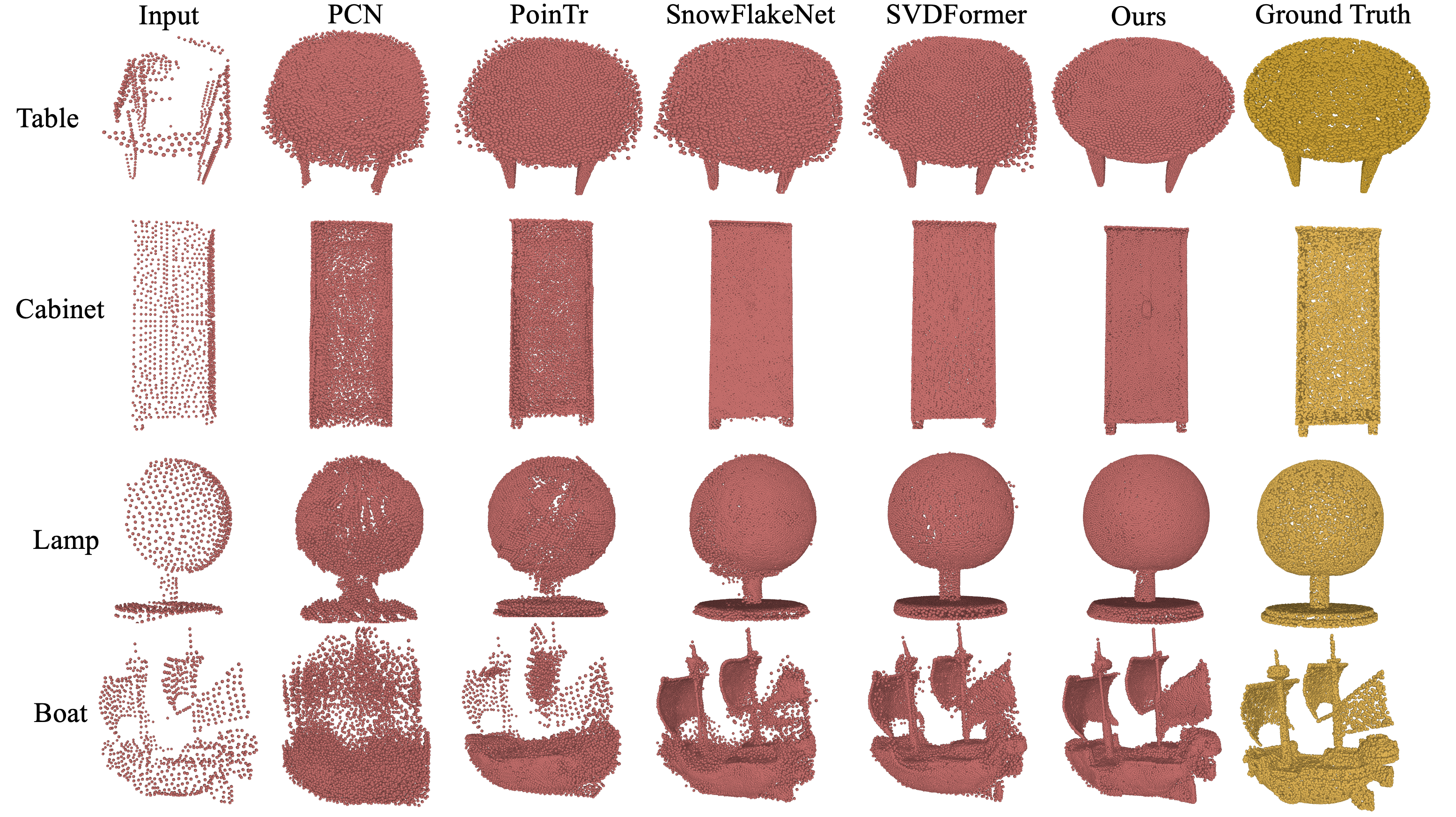}
  \caption{Visualizartion on the PCN Dataset.}
  \label{p3}
\end{figure}

\paragraph{Result on PCN}
Table \ref{PCN} summarizes the $CD_{l1}$ comparisons on eight categories of the PCN\cite{yuan2018pcn} dataset. VD-PCN consistently outperforms all baselines in terms of both per-category $CD_{l1}$  and the averaged distance. In general, lower Chamfer Distance indicates more accurate reconstructive shape. Specifically, our model achieves the averaged $CD_{l1}$ 6.32 which reduces the Chamfer Distance of the best competitor SVDFormer\cite{zhu2023svdformer} by 0.22. Though both of SVDFormer and VD-PCN rebuild the 3D shape with a multi-view projection way. SVDFormer utilize the combination of point information and image information while VD-PCN fully leverage the 2d information and the orderliness of 2d pixels. As indicated by the results, the knowledge transfer help our model predict better coarse point $P_c$. 

We also provide a qualitative analysis on PCN dataset. Figure \ref{p3} further visualizes the point cloud completion results of four rather difficult categories. We choose two easier categories and two harder categories.  In particular, for the table, VD-PCN is the only model predicting a round desktop, comparing with other arts. For cabinet and lamp, VD-PCN predicts fine-grained detail (i.e. the handle on the cabinet) and less noise. Also, VD-PCN performs better on the difficult boat, not only recovers more detail from limited input information but also brings less noise. We also provide visualization on other categories and failure case analysis in the supplementary material.

\paragraph{Result on ShapNet 55}
The test set of ShapeNet-55 can be classified into three levels of difficulty: simple (S), moderate (M), and hard (H), which correspond to different numbers of missing points (2,048, 4,096, and 6,144). Under all the difficulty, the the input points are subsample to 2048 considering the fixed width of a neural network. However, since we are processing the 2d images, we can accept more points than the point-based methods, which is a natural advantage, but for fairness, we still choose a common evaluation methods. The quantitative results are presented in table \ref{S55}. The results under different difficulty levels are shown in performance comparison in terms of L2 Chamfer Distance $\times10^3$ (CD$_{L2}$) on the ShapeNet-55 dataset. The per-category L2 Chamfer Distance results are reported on 5 categories with most training samples and 5 categories with the least training samples. 

\begin{table}[htbp]
\centering
\resizebox{\textwidth}{!}{
\begin{tabular}{@{}l|ccccc|ccccc|c@{}}
\toprule
Method & Table & Chair & Plane & Car & Sofa & Birdhouse & Bag & Remote & Keyboard & Rocket  & CD$_{L2}$  \\ 
\midrule
FoldingNet\cite{yang2018foldingnet}  & 2.53 & 2.81 & 1.43 & 1.98 & 2.48 & 4.71 & 2.79 & 1.44 & 1.24 & 1.48  & 3.12  \\
TopNet\cite{tchapmi2019topnet}  & 2.21 & 2.53 & 1.14 & 2.18 & 2.36 & 4.83 & 2.93 & 1.49 & 0.95 & 1.32  & 2.91  \\
PCN\cite{yuan2018pcn}  & 2.13 & 2.29 & 1.02 & 1.85 & 2.06 & 4.50 & 2.86 & 1.33 & 0.89 & 1.32  & 2.66  \\
GRNet\cite{xie2020grnet}  & 1.63 & 1.88 & 1.02 & 1.64 & 1.72 & 2.97 & 2.06 & 1.09 & 0.89 & 1.03  & 1.97  \\
PoinTr\cite{yu2021pointr} & 0.81 & 0.95 & 0.44 & 0.91 & 0.79 & 1.86 & 0.93 & 0.53 & 0.38 & 0.57  & 1.09  \\
SeedFormer\cite{zhou2022seedformer}& 0.72 & 0.81 & 0.40 & 0.89 & 0.71 & 1.51 & 0.79 & 0.46 & 0.36 & 0.50  & 0.92  \\
AnchorFormer\cite{chen2023anchorformer} & 0.58 & 0.67 & 0.33 & 0.69 & 0.58 & 1.35 & 0.64 & 0.36 & 0.27 & 0.42 & 0.76 \\
\textbf{Ours} & \textbf{0.52} & \textbf{0.60} & \textbf{0.39} & 0\textbf{0.62} & \textbf{0.49} & \textbf{1.15} & \textbf{0.53} & \textbf{0.33} & \textbf{0.24} & \textbf{0.41} & \textbf{0.70} \\
\bottomrule
\end{tabular}
}
\caption{Comparision on the ShapeNet 55 dataset}
\label{S55}
\end{table}

\paragraph{Result on MVP} We further test our model on the MVP dataset, The MVP dataset\cite{pan2021variational} is a multi-view partial point cloud dataset which consists of
16 categories of high-quality partial/complete point clouds.
For each complete CAD shape, 26 partial point clouds are generated by selecting 26 camera poses which are uniformly distributed on a unit sphere. The resolution of depth camera used MVP is higher to stimulate the real scene. We show a long table at 16384 resolution and provide other resolution at the supplementary material. Measure with $CD_{l2}$, our model out perform strong baseline FBNet. The result is listed in table \ref{mvp}.
\begin{table}[ht]
    \centering
    \resizebox{\textwidth}{!}{
    \begin{tabular}{c| c c c c c c c c c c c c c c c c |c}
        \hline
        \textbf{Methods} & \rotatebox{70}{\textbf{airplane}} & \rotatebox{70}{\textbf{cabinet}} & \rotatebox{70}{\textbf{car}} & \rotatebox{70}{\textbf{chair}} & \rotatebox{70}{\textbf{lamp}} & \rotatebox{70}{\textbf{sofa}} & \rotatebox{70}{\textbf{table}} & \rotatebox{70}{\textbf{watercraft}} & \rotatebox{70}{\textbf{bed}} & \rotatebox{70}{\textbf{bench}} & \rotatebox{70}{\textbf{bookshelf}} & \rotatebox{70}{\textbf{bus}} & \rotatebox{70}{\textbf{guitar}} & \rotatebox{70}{\textbf{motorbike}} & \rotatebox{70}{\textbf{pistol}} & \rotatebox{70}{\textbf{skateboard}} & \rotatebox{70}{\textbf{Avg.}} \\
        \hline
        PCN  & 2.95 & 4.13 & 3.04 & 7.07 & 14.93 & 5.56 & 7.06 & 6.08 & 12.72 & 5.73 & 6.91 & 2.46 & 1.02 & 3.53 & 3.28 & 2.99 & 6.02 \\
        TopNet  & 2.72 & 4.25 & 3.40 & 7.95 & 17.01 & 6.04 & 7.42 & 6.04 & 11.60 & 5.62 & 8.22 & 2.37 & 1.33 & 3.90 & 3.97 & 2.09 & 6.36 \\
        MSN  & 2.07 & 3.82 & 2.76 & 6.21 & 12.72 & 4.74 & 5.32 & 4.80 & 9.93 & 3.89 & 5.85 & 2.12 & 0.69 & 2.48 & 2.91 & 1.58 & 4.90 \\
        CRN  & 1.59 & 3.64 & 2.60 & 5.24 & 9.02 & 4.42 & 5.45 & 4.26 & 9.56 & 3.67 & 5.34 & 2.23 & 0.79 & 2.23 & 2.86 & 2.13 & 4.30 \\
        GRNet& 1.61 & 4.66 & 3.10 & 4.72 & 5.66 & 4.61 & 4.85 & 3.53 & 7.82 & 2.96 & 4.58 & 2.97 & 1.28 & 2.24 & 2.11 & 1.61 & 3.87 \\
        NSFA & 1.51 & 4.24 & 2.75 & 4.68 & 6.04 & 4.29 & 4.84 & 3.02 & 7.93 & 3.87 & 5.99 & 2.21 & 0.78 & 1.73 & 2.04 & 2.14 & 3.77 \\
        VRCNet & 1.15 & 3.20 & 2.14 & 3.58 & 5.57 & 3.58 & 4.17 & 2.47 & 6.90 & 2.76 & 3.45 & 1.78 & 0.59 & 1.52 & 1.83 & 1.57 & 3.06 \\
        SnowflakeNet & 0.96 & 3.19 & 2.27 & 3.30 & 4.10 & 3.11 & 3.43 & 2.29 & 5.93 & 2.29 & 3.34 & 1.81 & 0.50 & 1.72 & 1.54 & 2.13 & 2.73 \\
        FBNet  &  0.81  &  2.97  &  2.18  &  2.83  &  2.77  &  2.86  &  2.84  &  1.94  &  \textbf{4.81}  &  1.94  &  2.91  &  1.67  &  0.40  &  1.53  &  1.29  &  1.09  &  2.29  \\
        \textbf{Ours} & \textbf{0.65} & \textbf{2.82} & \textbf{2.11} & \textbf{2.58} & \textbf{2.74} & \textbf{2.69} & \textbf{2.72} & \textbf{1.82} & 4.82 & \textbf{1.65} & \textbf{2.85} & \textbf{1.55} & \textbf{0.34} & \textbf{1.38} & \textbf{1.24} & \textbf{0.96} & \textbf{2.16} \\
        \hline
    \end{tabular}
    }
    \caption{Point cloud completion results on MVP dataset (16384 points) in terms of per-point L2 Chamfer distance (×10$^4$).}
    \label{mvp}
\end{table}

\begin{table}[ht]
    \centering
        \begin{tabular}{cccccc}
        \toprule
        \textbf{Methods} & \textbf{$CD_{l2}-S$} & \textbf{$CD_{l2}-M$} & \textbf{$CD_{l2}-H$} & \textbf{$CD_{avg}$}  & \textbf{F-Score\% $\uparrow$} \\
        \midrule
        FoldingNet\cite{yuan2018pcn}           & 2.67 & 2.66 &4.05 &3.12 &0.082  \\
        TopNet\cite{tchapmi2019topnet}         & 2.26 & 2.16 &4.30 &2.91 &0.126 \\
        PCN\cite{liu2020morphing}              & 1.94 & 1.96 &4.08 &2.66 &0.133 \\
        GRNet\cite{wang2020cascaded}           & 1.35 & 1.71 &2.85 &1.97 &0.238 \\
        PoinTr\cite{pan2020ecg}                & 0.58 & 0.88 &1.79 &1.09 &0.464 \\
        SeedFormer\cite{pan2021variational}    & 0.50 & 0.77 &1.49 &0.92 &0.472 \\
        AnchorFormer\cite{chen2023anchorformer}&0.41 &0.61 &1.26   &0.76 &\textbf{0.558} \\
        SVDFormer\cite{zhu2023svdformer}       &0.48 &0.70 &1.30   &0.83 &0.451 \\
        CRA-PCN \cite{rong2024cra}             &0.48 &0.71 &1.37   &0.85 &- \\
        \textbf{Ours} & \textbf{0.36} & \textbf{0.56} & \textbf{1.19} & \textbf{0.70} &\textbf{0.558} \\
        \bottomrule
        \end{tabular}
        \caption{Comparison of methods on the MVP dataset based on CD-$\ell_2$ and F-Score.}
        \label{S3}
\end{table}

\subsection{Analysis of VD-PCN}
In this section, we do an accuracy-complexity trade-off test, evaluate the effectiveness of the distillation method and analyse different distillation strategy. We also discuss the motivation and current limitation of our work. 

\paragraph{Accuracy-Complexity Trade-Offs}We report the run-time FLOPs and inference time in Tab \ref{complex} and discuss the time consumption for the teacher training phase. All methods were evaluated on a single NVIDIA GeForce GTX 4090Ti graphic card. For fair comparisons, we disable gradient calculation and use point clouds with a resolution of 2,048. VD-PCN needs additional computation cost at teacher's training while the distillation process is faster than training from scratch. And the teacher neither appear at the inference time nor provide any information. From the results on table \ref{complex}, we see VD-PCN can achieve better trade-offs in performance and complexity. The time indicate the inference running time in unit input. 

\paragraph{Analysis of the different distillation strategy}Also we experiment with different distillation strategy. In the CNN pooling operation, feature map $F_v \in \mathbb{R}^{k \times C \times H_1 \times W_1}$ would be further max pooled into a gloval feature tensor $F_g \in \mathbb{R}^{k \times C }$. In the table \ref{distill}, we find that apply both $f_v$ and $f_g$ would produce better result, which also correspond with empirical analysis.

\begin{table}[ht]
    \centering
    
    \begin{minipage}{0.5\linewidth}
        \resizebox{\textwidth}{!}{
        \begin{tabular}{c|c|c|c}
            \toprule
            \textbf{Methods} & \textbf{FLOPs} (G) & \textbf{Time} (ms) & \textbf{$CD_{l1}$} \\
            \midrule 
            PoinTr\cite{yu2021pointr} & 18.81  & 15.34  & 8.38 \\
            SeedFormer\cite{zhou2022seedformer} & 107.51 & 16.50  & 6.74 \\
            AnchorFormer\cite{chen2023anchorformer} &\textbf{14.54} &23.71 &6.59 \\
            SVDFormer\cite{zhu2023svdformer} & 50.14 & 17.09  & 6.54 \\
            \textbf{Ours} & 105.49 & \textbf{15.00}  & \textbf{6.32} \\
            \bottomrule
        \end{tabular}
         }
        \caption{Run-time analysis for FLOPs and inference time, both are the smaller the better.}
        \label{complex}
    \end{minipage}%
    \hfill
    \centering
    \begin{minipage}{0.45\linewidth}
        \resizebox{\textwidth}{!}{
        \begin{tabular}{c|c|c|c|c}
            \toprule
            \textbf{Model} & \textbf{$f_{v}$ }  & \textbf{$f_{g}$ } & $CD_{l1}$ & \textbf{F-Score\% $\uparrow$} \\
            \midrule 
            A &           &             & 6.55          & 0.841 \\ 
            B &\checkmark &             & 6.38          & 0.850\\ 
            C &           & \checkmark  & 6.34          & 0.851\\ 
            D &\checkmark & \checkmark  & \textbf{6.32} & \textbf{0.852} \\ 
            \bottomrule
        \end{tabular}
        }
        \caption{Ablation studies on the distillation strategy. Model A do not apply the knowledge transfer and share the exact model with other variants.}
        \label{distill}
    \end{minipage}
\end{table}

\paragraph{Limitation}
The major limitation of our work is the lack of study of a real-world scenario for the proposed framework. All the common used dataset (i.e. PCN, ShapeNet 55, MVP) are artificially derived from ShapeNet\cite{chang2015shapenet}.  In future work, we will focus on extending the work to real acquisitions, such as point cloud scene and LiDAR point cloud.

\section{Conclusion}
In this work, we presented VD-PCN, a novel approach for 3D point cloud completion that synergistically combines multi-view 2D representations with a knowledge distillation strategy. By leveraging mature 2D CNN models and the orderliness of 2D pixel data, our method achieves superior performance over prior point-based methods on various benchmarks. The multi-view encoder provides a panoramic understanding of the partial point cloud, while the distillation from a well-trained teacher model guides the student towards accurate completions. The following dual-modality decoder that integrates both 2D and 3D information for upsampling the coarse point cloud predictions to high resolution. Extensive experiments demonstrate the superiority of our method. Moreover, It will be an interesting future direction to extend the knowledge transfer strategy to similar completion-like tasks.



\newpage
\bibliography{ref}
\bibliographystyle{unsrtnat}


\newpage

\section*{NeurIPS Paper Checklist}
\begin{enumerate}

\item {\bf Claims}
    \item[] Question: Do the main claims made in the abstract and introduction accurately reflect the paper's contributions and scope?
    \item[] Answer: \answerYes{} 
    \item[] Justification: In the abstract and introduction, we share two observation in the completion task and give out solution.

\item {\bf Limitations}
    \item[] Question: Does the paper discuss the limitations of the work performed by the authors?
    \item[] Answer: \answerYes{} 
    \item[] Justification: In the 4.3 Analysis of model part, we give an discussion about the limitation of current work.

\item {\bf Theory Assumptions and Proofs}
    \item[] Question: For each theoretical result, does the paper provide the full set of assumptions and a complete (and correct) proof?
    \item[] Answer: \answerNo{}
    \item[] Justification: There is no theoretical result in this paper. 

    \item {\bf Experimental Result Reproducibility}
    \item[] Question: Does the paper fully disclose all the information needed to reproduce the main experimental results of the paper to the extent that it affects the main claims and/or conclusions of the paper (regardless of whether the code and data are provided or not)?
    \item[] Answer: \answerYes{} 
    \item[] Justification: We would provide code, training detail, error bar, and model weight for Reproducibilty.

\item {\bf Open access to data and code}
    \item[] Question: Does the paper provide open access to the data and code, with sufficient instructions to faithfully reproduce the main experimental results, as described in supplemental material?
    \item[] Answer: \answerYes{} 
    \item[] Justification: Yes, we would provide open access.

\item {\bf Experimental Setting/Details}
    \item[] Question: Does the paper specify all the training and test details (e.g., data splits, hyperparameters, how they were chosen, type of optimizer, etc.) necessary to understand the results?
    \item[] Answer: \answerYes{} 
    \item[] Justification: We have sepcified all the training details in.

\item {\bf Experiment Statistical Significance}
    \item[] Question: Does the paper report error bars suitably and correctly defined or other appropriate information about the statistical significance of the experiments?
    \item[] Answer: \answerYes{} 
    \item[] Justification: We conduct multiple times of experiment and ablation study and provide the error bar to verify the statistical significance.

\item {\bf Experiments Compute Resources}
    \item[] Question: For each experiment, does the paper provide sufficient information on the computer resources (type of compute workers, memory, time of execution) needed to reproduce the experiments?
    \item[] Answer: \answerYes{} 
    \item[] Justification: For the experiments, we mainly finish it on a 4 * Nvidia A800 GPU. But 4* Nvidia 3090 GPU can reproduce the whole experiment. Training on the PCN dataset would cost 1.5day. 
    
\item {\bf Code Of Ethics}
    \item[] Question: Does the research conducted in the paper conform, in every respect, with the NeurIPS Code of Ethics \url{https://neurips.cc/public/EthicsGuidelines}?
    \item[] Answer: \answerYes{} 
    \item[] Justification: We have checked the NeurIPS code of Ethics and the whole research was  
    the research conducted in the paper conform with the NeurIPS Code of Ethics.

\item {\bf Broader Impacts}
    \item[] Question: Does the paper discuss both potential positive societal impacts and negative societal impacts of the work performed?
    \item[] Answer: \answerNA{} 
    \item[] Justification: This application paper do not talk about broader impacts. 
    
\item {\bf Safeguards}
    \item[] Question: Does the paper describe safeguards that have been put in place for responsible release of data or models that have a high risk for misuse (e.g., pretrained language models, image generators, or scraped datasets)?
    \item[] Answer: \answerNA{}  
    \item[] Justification: There is no societal impact of the work performed.

\item {\bf Licenses for existing assets}
    \item[] Question: Are the creators or original owners of assets (e.g., code, data, models), used in the paper, properly credited and are the license and terms of use explicitly mentioned and properly respected?
    \item[] Answer: \answerYes{} 
    \item[] Justification: The creators and original owners are properly credited.

\item {\bf New Assets}
    \item[] Question: Are new assets introduced in the paper well documented and is the documentation provided alongside the assets?
    \item[] Answer: \answerYes{} 
    \item[] Justification: We give our code, model, training detail and licence in the supplementary material and would provide it in a structured template later.

\item {\bf Crowdsourcing and Research with Human Subjects}
    \item[] Question: For crowdsourcing experiments and research with human subjects, does the paper include the full text of instructions given to participants and screenshots, if applicable, as well as details about compensation (if any)? 
    \item[] Answer: \answerNA{} 
    \item[] Justification: The paper does not involve crowdsourcing nor research with human subjects

\item {\bf Institutional Review Board (IRB) Approvals or Equivalent for Research with Human Subjects}
    \item[] Question: Does the paper describe potential risks incurred by study participants, whether such risks were disclosed to the subjects, and whether Institutional Review Board (IRB) approvals (or an equivalent approval/review based on the requirements of your country or institution) were obtained?
    \item[] Answer: \answerNA{} 
    \item[] Justification: The paper does not involve crowdsourcing nor research with human subjects.

\end{enumerate}

\end{document}